\documentclass[runningheads]{llncs}
\usepackage{graphicx}
\usepackage{float}
\usepackage{multirow}
\usepackage{unicode-math}

\begin{document}
\title{Deep Learning-Based Image Recognition for Soft-Shell Shrimp Classification}
\titlerunning{Soft-Shell Shrimp Classification}
%
\author{Yun Hao Zhang \inst{1}, I-Hsien Ting\inst{2}, Dario Liberona\inst{3}, Yun-Hsiu Liu\inst{4} and Kazunori Minetaki\inst{5}}
\authorrunning{Zhang et al.}
\institute{National University of Kaohsiung, Taiwan\\
\email{d07220256@gmail.com}\\
\and
National University of Kaohsiung, Taiwan\\
\email{iting@nuk.edu.tw}\\
\and
Seinäjoki University of Applied Sciences, Finland\\
\email{Dario.Liberona@seamk.fi}\\
\and
National University of Kaohsiung, Taiwan\\
\email{yvonneliu0616@gmail.com}\\
\and
Kindai University, Japan\\
\email{kminetaki@bus.kindai.ac.jp}\\
}
\maketitle              
\begin{abstract}
With the integration of information technology into aquaculture, production has become more stable and continues to grow annually. As consumer demand for high-quality aquatic products rises, freshness and appearance integrity are key concerns. In shrimp-based processed foods, freshness declines rapidly post-harvest, and soft-shell shrimp often suffer from head-body separation after cooking or freezing, affecting product appearance and consumer perception. To address these issues, this study leverages deep learning-based image recognition for automated classification of white shrimp immediately after harvest. A convolutional neural network (CNN) model replaces manual sorting, enhancing classification accuracy, efficiency, and consistency. By reducing processing time, this technology helps maintain freshness and ensures that shrimp transportation businesses meet customer demands more effectively.

\end{abstract}
\section{Introduction}
In the early days, the supply of shrimp for consumption mainly relied on wild capture as the primary source. According to a report released by the Food and Agriculture Organization of the United Nations (FAO) in 2022, the global production of farmed shrimp has been increasing steadily. As early as 2013, the quantity of aquaculture-produced seafood for human consumption had already surpassed the supply from marine capture and continued to grow annually. Notably, Asian countries accounted for 88\% of the world's total shrimp production. Today, aquaculture has largely replaced wild capture as the primary source of aquatic animal products for human consumption.

According to Taiwan’s Fisheries Research Institute under the Ministry of Agriculture, Taiwan began shrimp farming research as early as 1968. At its peak, Taiwan’s annual shrimp production reached 100,000 metric tons, which was significant compared to the world's total production of approximately 400,000 metric tons at the time, earning it the title of the "Shrimp Farming Kingdom." 

The process of harvesting and processing white shrimp from aquaculture ponds is complex, requiring multiple stages of sorting and selection to ensure freshness and extend shelf life. Special attention must also be given to the shrimp's appearance. Traditionally, the sorting process involves shrimp farmers harvesting the shrimp from ponds and selling them directly to intermediary shrimp transporters, who then handle the cleaning, sorting, and freezing procedures. During harvesting, white shrimp may suffer minor damage, or some shrimp may have unique growth characteristics that result in physical imperfections. Therefore, after being harvested and brought ashore, white shrimp must go through cold water cleaning and freezing, but the most critical step is the classification stage. This classification is done manually, with shrimp being sorted based on customer requirements for size. A particularly important classification criterion is the hardness of the shrimp’s shell, which is used to distinguish different grades.

One key reason for distinguishing shrimp based on shell hardness is the varying needs of customers regarding shrimp processing and usage. For most food service providers, shell hardness does not make a significant difference, as they select shrimp based on the type needed for their menu. However, for food processing companies, shell hardness is a crucial factor. These companies prioritize long-term preservation and product appearance. When frozen, soft-shell shrimp tend to develop dents and deformities, significantly reducing their visual appeal. Additionally, when cooked—either during processing or by consumers at home—soft-shell shrimp are less able to protect the shrimp body, making them prone to issues such as the shrimp head detaching from the body or the shell sticking to the meat, making peeling difficult. These factors affect not only the product's visual appeal and consumer willingness to purchase but may also lead consumers to mistakenly perceive the product as inferior, raising concerns about its quality.

The shell hardness of shrimp varies over time, as shrimp undergo molting at irregular intervals. After molting, it takes time for their soft shell to harden again. This makes classification difficult, as shrimp caught during this transition period fall into a gray area that is hard to distinguish. Moreover, individual sorting experts may have differing opinions when evaluating shrimp, leading to inconsistencies in classification standards. This, in turn, affects the ability to meet customer requirements and indirectly impacts product presentation and consumer acceptance.

Based on the issues outlined above, this study intends to integrate AI technology with the shrimp farming industry, focusing on white shrimp as the primary research subject. To address the challenges faced by shrimp transporters in the classification stage, this study collects images of live white shrimp, capturing detailed visual characteristics of soft-shell and hard-shell shrimp. A convolutional neural network (CNN) is used for model training to apply image recognition technology as a substitute for traditional manual sorting.

This approach aims to improve consistency in the classification of soft-shell and hard-shell shrimp to better meet customer demands and reduce the likelihood of dissatisfaction due to mismatched product expectations. Moreover, we would like to enhance efficiency by significantly reducing the time required for shrimp sorting, shortening the time from harvest to customer delivery, and maintaining optimal freshness for consumers. Thus, this research investigates the use of deep learning technology for identifying soft-shell and hard-shell shrimp through image recognition.

\section{Literature Review}
\subsection{Differences Between Soft-Shell Shrimp and Hard-Shell Shrimp}

\begin{figure}[H]
\centering
\includegraphics[width=0.6\textwidth]{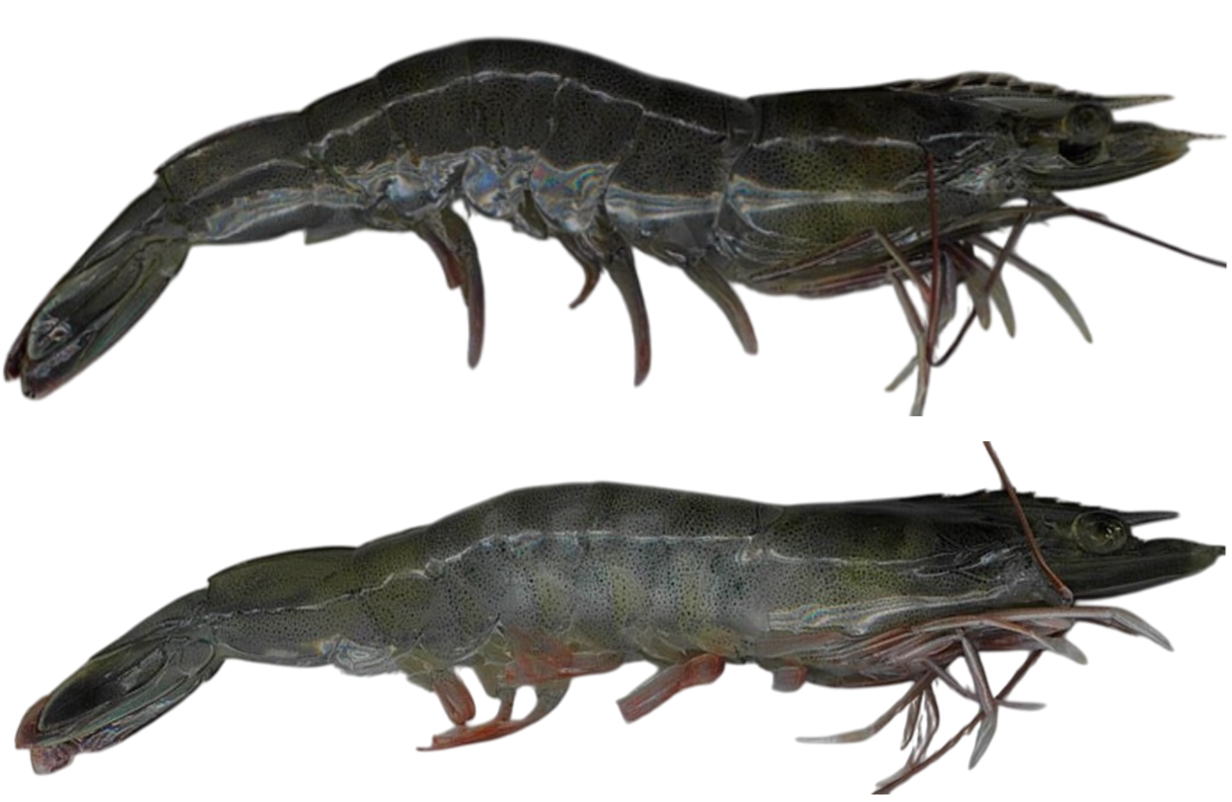}
\caption{Comparison of Hard-Shell Shrimp (Top) and Soft-Shell Shrimp (Bottom)}\label{fig1}
\end{figure}

The occurrence of the soft-shell phenomenon is primarily influenced by protein and amino acid content. Simply put, after molting, white shrimp consume excessive nutrients, leading to nutritional deficiencies or imbalances that prevent the normal hardening of the new shell. The main visual characteristic of soft-shell shrimp is that the shell clings tightly to the shrimp’s flesh, with visible wrinkles and indentations on the surface. As shown in Figure \ref{fig1} , some soft-shell shrimp may appear less visually distinct from hard-shell shrimp due to a longer recovery period after molting. In these cases, the shell may have regained its original shape but has not yet fully hardened, making it difficult to differentiate from hard-shell shrimp.

Tactile characteristics can be identified by gently pressing the shell on the shrimp’s head. If the shell easily sinks inward, it is a soft-shell shrimp; in contrast, a hard-shell shrimp will remain firm. While molting allows shrimp to develop a larger shell to accommodate their growing bodies, they are at their most vulnerable state during the transition period before the new shell hardens. Without the protection of a hardened exoskeleton, shrimp are more susceptible to environmental factors that can cause physical damage or even death.

Each shrimp’s molting cycle varies, resulting in differences in shell-hardening timelines. Therefore, when shrimp farmers harvest white shrimp, some shrimp may still be in the soft-shell stage. Since shrimp farmers cannot control the exact timing of molting, they can only try to avoid harvesting during peak molting periods to minimize post-harvest losses caused by soft-shell shrimp.

\subsection{Aquaculture Image Recognition Technology}
With the continuous advancement of information technology, the integration of IT with traditional industries is becoming increasingly common. The introduction of information technology has greatly benefited traditional industries, leading to significant improvements in both quality and production levels compared to the past. Traditionally, aquaculture relied heavily on manual farming, which had limitations in terms of labor capacity. Managing an extensive farming area manually made it difficult to oversee all aspects effectively, resulting in inconsistent production volumes and challenges in maintaining high-quality standards.

By leveraging information technology, system models are gradually replacing human labor in handling simpler tasks, alleviating workload pressures on aquaculture farmers. According to research by Zhao et al., the implementation of image recognition systems has been used to assist in measuring the individual length of fish in aquaculture ponds. By using an overhead imaging approach, underwater fish sizes can be detected and analyzed, providing corresponding data while addressing overlapping image issues caused by fish movement. This significantly reduces labor and time requirements for aquaculture farmers while minimizing disturbances to the feeding environment and behavior of the fish \cite{b1}.

A study by Silva et al. focused on Litopenaeus vannamei (Pacific white shrimp) and applied Passive Acoustic Monitoring (PAM) to capture the biting sounds produced by shrimp during feeding. These sound frequencies were converted into visual waveform and spectrogram representations to monitor shrimp feeding intensity and behavior. The frequency of these sounds serves as an indicator of feed consumption, allowing for an accurate assessment of shrimp feeding levels \cite{b2}.

Building on Silva’s research, Du et al. extended these methods to fish farming, incorporating a fusion approach that combines both acoustic and imaging technologies. By capturing the biting sounds of fish during feeding and converting them into color-coded frequency maps, while simultaneously recording images of their feeding state, researchers could filter out non-feeding noises. They developed an enhanced LC-GhostNet lightweight network structure to improve classification accuracy in evaluating fish feeding intensity. This advancement helps optimize feed distribution, reducing excess feed that may pollute water quality and decreasing overall aquaculture costs \cite{b3}.

A study by Strebel et al. also utilized Passive Acoustic Monitoring (PAM) to detect the feeding sounds of Pacific white shrimp and integrated this data into an automated feeding system. The system dynamically adjusted feed distribution based on detected feeding frequencies, enabling multiple controlled feeding sessions per day while considering shrimp digestion time. This approach effectively reduced water pollution caused by excessive feed exposure and established a more efficient feed management strategy \cite{b4} \cite{b5} \cite{b6} \cite{b7}.

Water quality fluctuations often lead to disease outbreaks. To meet growing market demand, many aquaculture farms rely on intensive farming systems, which allow for high production volumes. However, high-density farming also increases the risk of disease transmission. Chen et al. studied high-density farming of grouper fish, highlighting how poor water quality can lead to higher disease susceptibility. Grouper fish exhibit visible physical abnormalities when infected, allowing researchers to detect diseases early. Their study proposed a two-stage ImageNet pre-trained deep learning model using a Convolutional Neural Network (CNN) to classify and identify three types of physical abnormalities in grouper fish. This method enables early disease detection, allowing for immediate intervention to reduce infection risks \cite{b8}.

Sladojevic et al. applied CNN-based deep learning for plant disease detection. Their study captured visual features of diseased plant leaves, labeled the diseases, and trained a model to recognize them. After training, the model successfully classified 13 different plant diseases and was able to distinguish healthy plants. This research aids farmers in early disease detection, minimizing crop loss and preventing disease spread \cite{b9}.

Jiang et al. developed an innovative freshness detection method using Colorimetric Sensor Arrays (CSA), which detect volatile organic compounds (VOCs) released by seafood. These chemical emissions are converted into visual images, and a CNN-based model is trained to recognize color changes, allowing for real-time seafood freshness monitoring \cite{b15}.

Liu et al. addressed a critical processing challenge: soft-shell shrimp, which can compromise product appearance and reduce consumer purchase intent. Their study collected shrimp samples from supermarkets and developed a deep learning model called Deep-ShrimpNet by modifying AlexNet’s architecture. After training, the model accurately classified soft-shell and hard-shell shrimp using image recognition. A combination of single and ensemble classifiers was used to enhance classification accuracy, ensuring that defective shrimp were effectively removed from the supply chain \cite{b16}.

Wang et al. conducted research on frozen red shrimp, collecting image data to simulate how freshness declines over time during transportation. They developed the ShrimpNet model to classify freshness levels in frozen shrimp and applied the Grad-CAM visualization method to generate heatmaps representing freshness variations. This approach ensures that frozen seafood remains safe and consumable after transportation, addressing consumer concerns about seafood safety \cite{b17}.

\section{The Research Method}
\subsection{The Research Architecture}

\begin{figure}[H]
\centering
\includegraphics[width=0.6\textwidth]{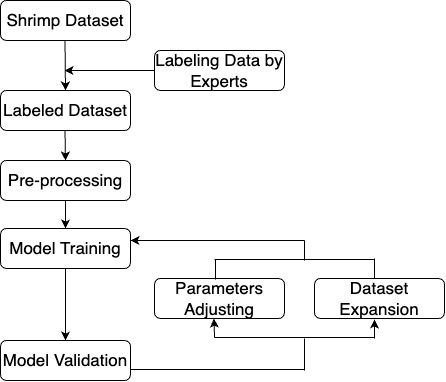}
\caption{The Research Architecture} \label{fig2}
\end{figure}

The research framework of this study is shown in Figure \ref{fig2}. This study obtained live white shrimp through collaboration with white shrimp aquaculture experts. First, shrimp classification experts manually distinguished between soft-shell shrimp and hard-shell shrimp. Next, the classified live white shrimp were scanned individually using the ApeosPort C3070 scanner to capture detailed physical characteristics, forming the necessary research dataset. Finally, the collected dataset was categorized into ordinary shrimp and soft-shell shrimp.

Before training the model, image preprocessing was applied to the white shrimp dataset to minimize the impact of irrelevant variables that could affect model performance. After preprocessing, the dataset entered the model training phase, where this study used Convolutional Neural Networks (CNNs) for training.

Once the model was trained, it underwent model validation, using overall accuracy as the key performance metric. If the accuracy was unsatisfactory, adjustments were made by either augmenting the dataset or tuning model parameters. After making the necessary modifications, the model was retrained and revalidated until optimal results were achieved, ensuring that the final model had a reliable classification capability.

\subsection{Data Source}
This study utilized white shrimp provided by collaborating shrimp aquaculture experts. After being harvested and brought ashore, the shrimp were first classified into soft-shell and hard-shell categories by experts. To ensure data accuracy, image scanning and feature capture were conducted while the shrimp were still alive, minimizing potential deterioration. Compared to the traditional process where shrimp transporters deliver harvested shrimp to processing plants for immediate washing and classification, this method ensures that the collected dataset better reflects the actual environmental conditions at the time of capture. To meet research requirements, the original dataset was split using the holdout method at a fixed ratio: Training set: 80\% of the data and Validation set: 20\% of the data. This dataset division was applied to facilitate model training and validation for the study.

\subsection{Data Pre-processing}

\begin{figure}[H]
\centering
\includegraphics[width=0.6\textwidth]{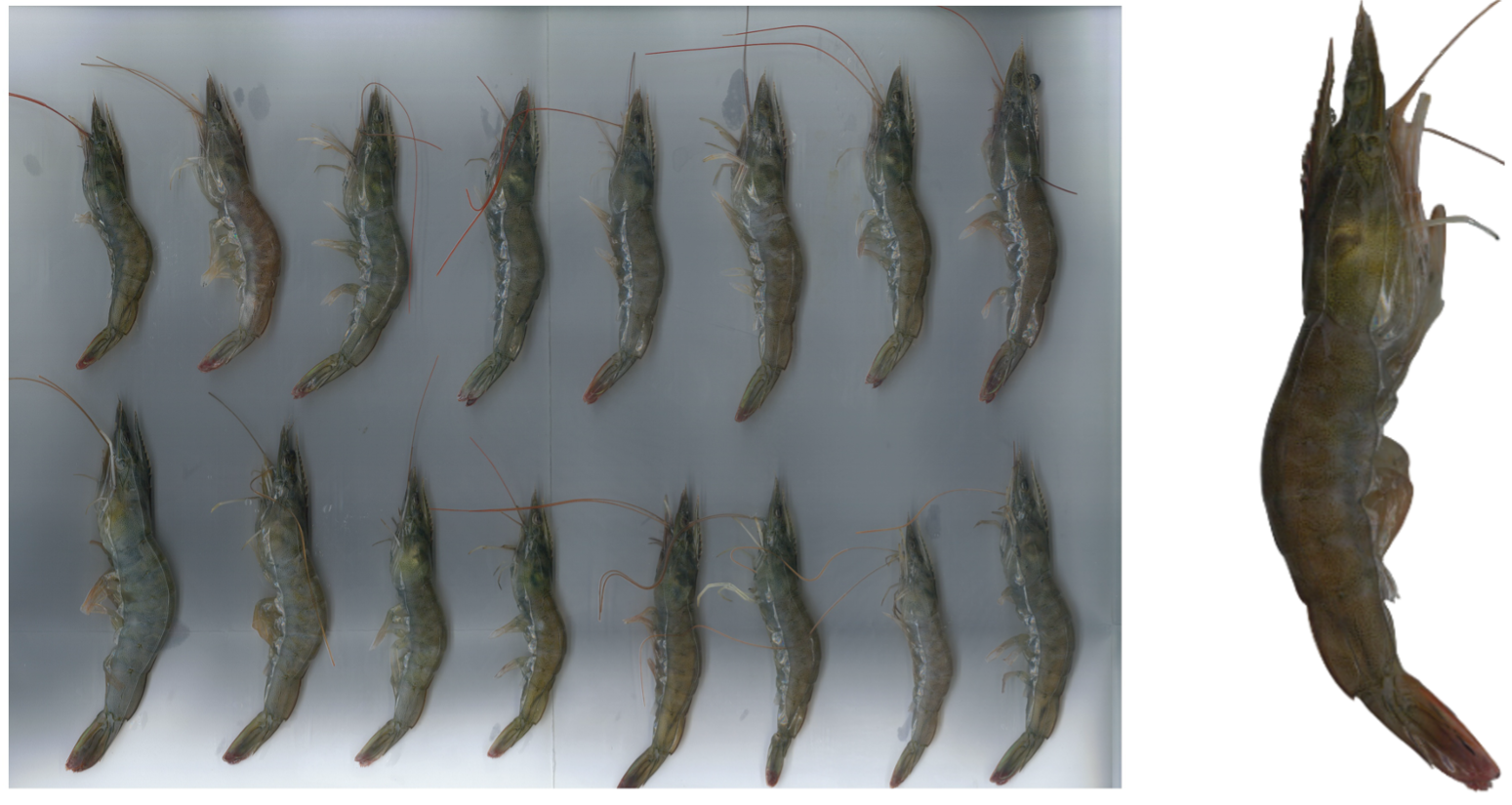}
\caption{The Process of Risk Assessment and Financial Analysis} \label{fig3}
\end{figure}

After the experts classified the white shrimp into soft-shell and hard-shell categories, the scanner was used to directly scan and capture dataset images, and the files were placed in the original folders of "ordinary-shrimp" and "soft-shell shrimp" respectively. As shown in the left-side image of Figure \ref{fig3}, in order to complete the collection of characteristic images within the survival time of the white shrimp, multiple shrimp were present in a single image. During the scanning process, due to the moisture released from the shrimp's body after leaving the water, some parts of the image background were stained with water. These issues may all contribute to errors in system judgment. To reduce the possibility of unrelated features, as shown in the right-side image of Figure \ref{fig3}, each image was cropped so that only one complete white shrimp remained, and the background was removed, leaving only the shrimp itself. This helps the model focus on extracting and learning valid features, thereby reducing the risk of influence from other factors as much as possible.

As shown in Figure \ref{fig3}, the dataset in this study was scanned using the ApeosPort C3070 scanner, with the original scanned image size being 9930×7020 pixels, the scanning resolution set to 600 dpi, and the scanning mode set to color scanning. To meet the research needs, the images were cropped, with the final image size ranging from 830×1280 pixels to 240×1280 pixels, for model training.

\subsection{The Architecture of The Training Model}
This study adopts Convolutional Neural Networks (CNN) for deep learning training. The concept of CNN is similar to the cognitive learning process of humans. When identifying an image, humans observe distinct points or lines in terms of color and then gradually form different shapes, using this abstraction method to establish a model. CNN mainly consists of convolutional layers, pooling layers, and fully connected layers.

First, the primary function of the convolutional layer is to retain the spatial structure of the image and extract features from it. In other words, each image is processed in a block-wise manner for feature extraction, generating a feature map. These blocks move across the image within a specific range, capturing features until the entire image is covered.

Second, the pooling layer functions to reduce parameters by subsampling the input patterns, thereby lowering computational costs. 

Finally, the fully connected layer acts as the final classifier. It integrates the previously extracted features through weight computation and determines the category to which the input image belongs. This process is similar to how humans identify individuals based on distinguishing features. As the last layer of CNN, the fully connected layer is crucial in this study because it combines features to ensure the model achieves effective recognition.

During the training and validation process, this study additionally incorporated the RMSProp optimizer. In practical applications, the loss function is often neither stable nor simple, and the corresponding error surface is usually complex and variable. Even within the same dimension, the learning rate may need rapid adjustments to accommodate different situations. RMSProp introduces a parameter to balance the influence of past and current gradients during learning rate adjustments. When the $\alpha$ value increases, past gradients are given more weight, meaning the adjustment process relies more on historical gradient information.

\subsection{Model Validation}
This study utilizes Convolutional Neural Networks (CNN) for model training. After training the model, the overall model accuracy is used as the primary criterion for model validation, with this value determining the feasibility of the model. During the model training process, the recorded values include the training accuracy (validation\_accuracy) and the sparse categorical cross-entropy loss function (sparse\_categorical\_crossentropy) from the cross-entropy-related loss function. Since model validation is conducted simultaneously, the recorded values also include validation accuracy and sparse categorical cross-entropy.

The loss function measures the difference between the model’s predicted values and the actual target values. Once the model calculates the loss function, it determines its loss value (Loss, inaccuracy). A higher loss value indicates lower accuracy, while a lower loss value suggests higher accuracy. Cross-entropy (cross-entropy) is used as the loss function for training the model in this study. There are two types of cross-entropy-related loss functions: categorical cross-entropy and sparse categorical cross-entropy. Both are used to compute the cross-entropy loss between the model’s predicted results and the actual labels in classification problems. The primary difference is that "sparse\_categorical\_crossentropy" is more convenient, as labels can be represented directly as integers without requiring one-hot encoding, whereas "categorical\_crossentropy" requires labels to be in one-hot encoded format. This study adopts sparse categorical cross-entropy as the loss function for model training.

After conducting the model study, this research presents the training data for each image during the training process, as well as the data for each validation process. To facilitate observation of fluctuations in overall model training and validation, the data from the process is visualized in curve charts. The numerical data is converted into graphical representations and displayed alongside the raw data, making it easier for researchers to analyze the model’s performance.

\section{Analysis Results}

This study was conducted using the Google Colaboratory platform for executing the research program. The required dataset was uploaded to Google Drive, and through synchronization between Google Colaboratory and Google Drive, the dataset was applied to the research program for execution. The study utilized Convolutional Neural Network (CNN) technology to train the model and recorded the data from the training process, presenting it in visualized charts. After training the model, validation was conducted immediately to determine the overall model accuracy, as well as the validation data and curve charts. The accuracy results were used to assess whether further evaluation of other model performance metrics should be carried out.
\begin{figure}[H]
\centering
\includegraphics[width=0.8\textwidth]{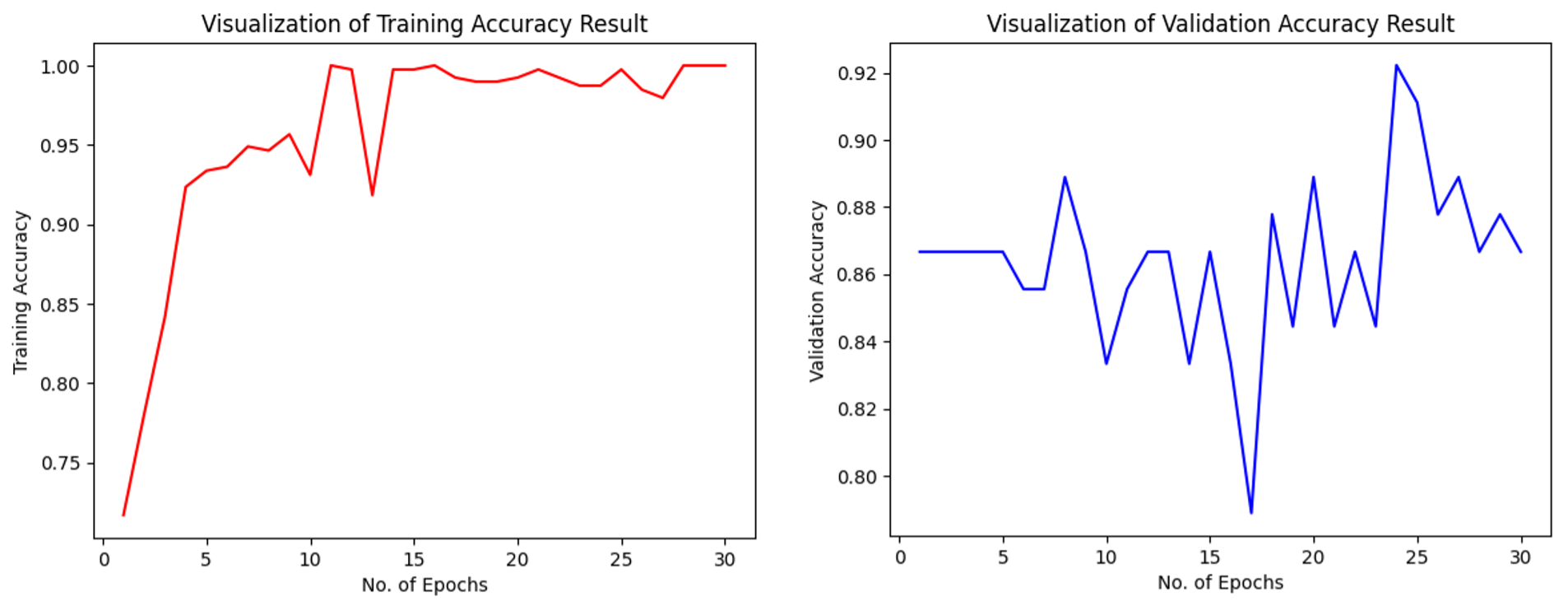}
\caption{0131Dataset-7 Training Curve Chart (Left) and Validation Curve Chart (Right)} \label{fig6}
\end{figure}
A total of 10 rounds of model training and validation were conducted in this study. Throughout the process, data augmentation and model parameter adjustments were continuously applied to achieve the optimal model performance. The dataset "0131Dataset-7" comprises images from the "1109dataset" and "1225dataset." The model validation accuracy reached approximately 87\%, surpassing the target accuracy set for the study. Compared to previous model validation results, the accuracy significantly improved, reaching the preset target value. Therefore, further evaluation of other model performance metrics was required. The training and validation results were visualized in curve charts, as shown in Figure 4. Based on these results, it can be inferred that, compared to previous data proportions, the proportion of training data for hard-shell shrimp needs to be increased to improve model training effectiveness and further enhance the overall model accuracy.

\begin{table}[!htbp]
\caption{\label{tbl1} The 0131Dataset-7 Dataset Summary Table}
\centering 
\begin{tabular}{|l|l|}
\hline
Dataset           & 0131Dataset-7 \\ \hline
Ordinary Shrimp   & 398           \\ \hline
Soft Shell Shrimp & 84            \\ \hline
Ratio             & 8:2           \\ \hline
Training Epoch    & 30            \\ \hline
Accuracy          & 87\%          \\ \hline
\end{tabular}
\end{table}

After completing the model training, validation was performed to determine the accuracy of the trained model using the dataset "0131Dataset-7." The resulting model accuracy was 86.66666746139526\%. The dataset contained 398 images of ordinary shrimp and 84 images of soft-shell shrimp used for model training. The training process consisted of 30 epochs, with a total training time of approximately 7 to 8 minutes. The overall model accuracy exceeded the study’s target of 85

This study used a total of 8 different dataset combinations for model training. Due to the limited number of soft-shell shrimp samples, collecting image data for them was also relatively difficult. Therefore, before training the model with these datasets, minor data augmentation was performed. Specifically, images with backgrounds removed were mirrored by flipping them horizontally. This was done to meet the minimum required number of samples, while avoiding excessive use of duplicate images due to human intervention, which could potentially invalidate the results.

\section{Conclusion and Future Research Suggestions}
The main concept of this study is to leverage information technology to collaborate effectively and practically with traditional industries, aiming to improve production and quality. A key emphasis of this research is the simulation of real-world environments to collect data images according to research needs, ensuring that the trained model can be applied in practical settings. To achieve this, live white shrimp were scanned and photographed, replicating the process where shrimp farmers harvest shrimp, which are then transported, washed, and sorted by shrimp transporters. Since the shrimp were still alive during scanning, the obtained accuracy results are closely aligned with real-world applications.

By utilizing Convolutional Neural Networks (CNN) to train the model, this study experimented with different dataset combinations and parameter adjustments to maximize recognition accuracy. Ultimately, through model validation, the highest accuracy achieved was approximately 87\%. The validation results confirm that the model possesses sufficient image classification capability and can directly distinguish soft-shell shrimp.

For future research suggestions, A collaborative strategy incorporating multiple classifiers could enhance model accuracy. According to relevant literature, combining CNN and SVM has been shown to significantly improve classification performance \cite{b18}. Implementing dual classification techniques with well-defined roles may lead to better classification outcomes. Additionally, the dataset should be continuously expanded to allow greater flexibility in model training and application. Particular attention should be given to balancing the ratio of ordinary shrimp and soft-shell shrimp in the dataset. A key insight from this study is the importance of data proportion, meaning that instead of focusing solely on increasing sample size, it is crucial to maintain a minimum of 30 samples while carefully adjusting the ratio between the two categories. Furthermore, improving the scanner’s resolution can help capture more refined details, enabling the model to extract additional distinguishing features. This would provide stronger classification criteria and improve overall recognition accuracy.

%
%
%
%

\end{document}